\documentclass[runningheads]{llncs}

\usepackage{helvet,times,courier}
\usepackage[utf8]{inputenc}      
\usepackage[T1]{fontenc}         
\usepackage[final]{microtype}    
\usepackage{amsmath,amsfonts}    
\usepackage{xspace}              
\usepackage{tikz}                
\usepackage[
  ruled
, vlined
, linesnumbered]{algorithm2e}    
\usepackage{todonotes}           

\usetikzlibrary{shapes,arrows,shadows,positioning,calc}
\tikzstyle{block} = [draw, rectangle, minimum height=1.8em, minimum width=9em, node distance=3.5em, transform shape]

\DontPrintSemicolon
\SetKwInput{Input}{input}

\newcommand{\refsec}[1]{Section~\ref{#1}}
\newcommand{\refdef}[1]{Definition~\ref{#1}}
\newcommand{\reffig}[1]{Fig.~\ref{#1}}

\newcommand{\wasp}{\textsc{wasp}\xspace}
\newcommand{\dwasp}{\textsc{dwasp}\xspace}
\newcommand{\dwaspgui}{\textsc{dwasp-gui}\xspace}
\newcommand{\gringo}{\textsc{gringo}\xspace}
\newcommand{\gringowrapper}{\textsc{gringo-wrapper}\xspace}
\newcommand{\aspide}{\textsc{aspide}\xspace}

\newcommand{\pr}{\Pi}                        
\newcommand{\dpr}[1]{\Delta_{#1}}            
\newcommand{\gpr}{\pr^G}                     
\newcommand{\gdpr}[1]{\Delta_{#1}^G}            
\newcommand{\bg}{\mathcal{B}}                
\newcommand{\debug}[2]{\_debug(#1,#2)}       
\newcommand{\support}[1]{\_support(#1)}      

\newcommand{\dnot}{\sim\!}                   
\newcommand{\imp}{\leftarrow}                

\title{An integrated Graphical User Interface for Debugging Answer Set Programs}
\author{Philip Gasteiger\inst{1}%
   \and Carmine Dodaro\inst{2}%
   \and Benjamin Musitsch\inst{1}%
   \and Kristian Reale\inst{2}%
   \and Francesco Ricca\inst{2}%
   \and Konstantin Schekotihin\inst{1}%
}
\authorrunning{P. Gasteiger et al.}
\institute{Alpen-Adria-Universität Klagenfurt, 9020 Klagenfurt, AT\\%
	       \email{\{firstname.lastname\}@aau.at}%
     \and  Universit\`a della Calabria, Rende CS, IT\\%
           \email{\{lastname\}@mat.unical.it}
}

\begin{document}
\maketitle

\begin{abstract}
Answer Set Programming (ASP) is an expressive knowledge representation and reasoning framework. 
Due to its rather simple syntax paired with high-performance solvers, ASP is interesting for industrial applications.
However, to err is human and thus debugging is an important activity during the development process.
Therefore, tools for debugging non-ground answer set programs are needed.
In this paper, we present a new graphical debugging interface for non-ground answer set programs.
The tool is based on the recently-introduced \dwasp approach for debugging and it simplifies the interaction with the debugger.
Furthermore, the debugging interface is integrated in \aspide, a rich IDE for answer set programs.
With our extension \aspide turns into a full-fledged IDE by offering debugging support.
\end{abstract}


\section{Introduction}
\label{sec:introduction}

Answer Set Programming (ASP)~\cite{DBLP:journals/cacm/BrewkaET11} is a declarative programming paradigm proposed in the area of logic programming and non-monotonic reasoning. 
Computational problems of comparatively high complexity can be modeled in the expressive language of ASP~\cite{DBLP:journals/tods/EiterGM97}, which provides a clear separation between the specification of a problem and the computation of its solutions by an ASP solver.
The rather simple syntax of ASP paired with high-performance solvers makes ASP a valuable tool for developing complex research and industrial applications~\cite{DBLP:conf/cpaior/AschingerDFGJRT11,DBLP:journals/ai/CalimeriGMR16,DBLP:conf/rr/DodaroLNR15,DBLP:conf/birthday/GrassoLMR11}.
Especially real-world applications outlined the advantages of ASP from a software engineering viewpoint. Namely, ASP programs are flexible, intuitive, extensible and easy to maintain~\cite{DBLP:conf/birthday/GrassoLMR11}.

Although the basic syntax of ASP is not particularly difficult, one of the most tedious and time-consuming  programming tasks is the identification of (even trivial) faults in a program. 
For this reason, several methodologies and tools have been proposed in the last few years for debugging ASP programs~\cite{Brain2005,Gebser2008b,Oetsch2010a,Oetsch2011step,Pontelli2009}, with the goal of making the development process faster and more comfortable.
 
We have recently proposed a new debugging technique in~\cite{dodaro:2015:interactive} that can be applied to non-ground ASP programs, and that allows to single out the rules causing a bug. 
This new approach overcomes the limits~\cite{Oetsch2010a} of state-of-the-art debuggers based on meta-programming~\cite{Gebser2008b,Oetsch2010a}, which suffer from a grounding blow-up problem.
Nonetheless the new technique was implemented a as a command line tool only, called \dwasp, that extends the \wasp solver~\cite{alviano:2015:advances}.

It is nowadays recognized that the development of programs can be made easier by development tools and graphic environments.
Indeed, the most diffused programming languages always come with the support of graphical debugging tools that are integrated in rich IDEs. 
As an example consider one of the most diffused debugging tools for C++, called \textit{gdb}.
Despite \textit{gdb} being shipped with \textit{g++} as a command line tool, given the complex nature of debugging, serious program inspections are often done by means of user friendly graphical tools (provided by IDEs such as Eclipse or Netbeans) that wrap the \textit{gdb} command.
Following this trend, the debugging approaches for ASP have been usually integrated in programming environment such as \aspide~\cite{febbraro:2011:aspide} or SeaLion~\cite{busoniu:2013:sealion,oetsch:2010:catching}.
However, initially \dwasp was not integrated in a graphic environment, and also \aspide featured only a limited support for debugging, which was restricted to ground programs.
In this paper we provide two contributions in this context:
\begin{enumerate}
\item A graphical user interface, called \dwaspgui, for \dwasp that improves the user-experience of the debugger.

\item A plug-in connector for \aspide, which integrates \dwaspgui within the IDE.
\end{enumerate}

The graphical user interface provides a more intuitive user-experience during the debugging process with \dwasp.
Furthermore, the integration in \aspide brings additional advantages to the users of \dwasp.
Indeed \aspide provides a unit-testing framework~\cite{febbraro:2011:unit} that was connected with \dwasp to automatically generate failing test cases for the debugger.
The integration is thus synergistic, as it simplifies the usage of the debugger and turns \aspide into a full-fledged IDE by offering a more advanced debugging tool.

\section{Answer Set Programming}
\label{sec:preliminaries}
In this section we recall the syntax and semantics of answer set programming.
Furthermore, some properties of answer set programs that are required for our debugging methodology are presented briefly.

\subsection{Syntax.}
A \emph{disjunctive logic program} (DLP) $\pr$ is a finite set of rules of the form
\begin{equation}
  a_1 \lor \ldots \lor a_m \imp l_1, \ldots, l_n
  \label{eq:rule}
\end{equation}
where $a_1, \ldots, a_m$ are atoms and $l_1, \ldots, l_n$ are literals for $m, n \geq 0$.
A \emph{literal} is an atom $a_i$ (positive) or its negation $\dnot a_i$ (negative), where $\dnot$ denotes the \emph{negation as failure}.
The complement of a literal $l$ and a set of literals $L$ is denoted by $\overline{l}$ and $\overline{L} := \{ \overline{l} \mid l \in L \}$, respectively, where $\overline{a} = \dnot a$ and $\overline{\dnot a} = a$ for an atom $a$.
An \emph{atom} is an expression of the form $p(t_1, \ldots, t_k)$, where $p$ is a predicate symbol and $t_1, \ldots, t_k$ are \emph{terms}, i.e. variables or constants.
An atom, literal, or rule is called \emph{ground}, if it is variable-free.
Given a rule $r$ of the form \eqref{eq:rule}, the set of atoms $H(r) = \{ a_1, \ldots, a_m \}$ is called \emph{head} and the set of literals $B(r) = \{ l_1, \ldots, l_n \}$ is called \emph{body}. Moreover, $B(r)$ can be partitioned into the sets $B^+(r)$ and $B^-(r)$ comprising the positive and negative body literals, respectively.
A rule $r$ is called \emph{fact} if $|H(r)| = 1$ and $B(r) = \emptyset$; \emph{constraint} if $H(r) = \emptyset$; and \emph{normal rule} if $|H(r)| = 1$ and $B(r) \neq \emptyset$.
For a fact $a \imp$ we omit the $\imp$ symbol and write $a$ instead.
Every rule $r \in \pr$ must be \emph{safe}, i.e. each variable of $r$ must occur in at least one positive literal of $B^+(r)$.

\subsection{Semantics.}
Let $\pr$ be an ASP program, $U_\pr$ be the Herbrand universe and $B_\pr$ be the Herbrand base of $\pr$.
Let $\gpr$ be the \emph{ground instantiation} of $\pr$ that is obtained by substituting variables with elements of $U_\pr$.
An \emph{interpretation} is a set of ground atoms $I \subseteq B_\pr$.
Given an interpretation $I$ a positive literal $l$ (its complement $\overline{l}$) is \emph{true} in $I$ iff $l\in I$ ($l\not\in I$).
An interpretation $M$ is a \emph{model} for $\gpr$ if for each rule $r \in \gpr$ having $B(r) \subseteq M$ it holds that $H(r) \cap M \neq \emptyset$.
Let $I_1$ and $I_2$ be two interpretations, then $I_1 \subseteq^+ I_2$ if and only if for each atom $a \in I_1$ it holds that $a \in I_2$.

Given the ground instantiation $\gpr$ of a DLP $\pr$ and an interpretation $I$, a \emph{reduct} of $\gpr$ w.r.t. $I$ is a ground program $\gpr_I$ obtained from $\gpr$ by: (i) deleting all rules $r \in \gpr$ whose negative body is false w.r.t to $I$ and (ii) deleting the negative body from the remaining rules.
An \emph{answer set} of $\pr$ is a model $M$ of $\gpr$ that is a $\subseteq^+$-minimal model of $\gpr_M$. 
Given the set of answer sets $AS(\pr)$ of $\pr$, the program $\pr$ is called \emph{incoherent}, if $AS(\pr) = \emptyset$, and \emph{coherent} otherwise.
\section{Debugging Approach}
\label{sec:debugging-approach}
In this section we present the debugging approach proposed in \cite{dodaro:2015:interactive} which is implemented inside the ASP solver \wasp \cite{alviano:2015:advances}.
First, the key idea behind the approach is presented on an abstract level in \refsec{sec:debugging-approach:idea}.
Afterwards, a way to integrate the approach inside an ASP solver is outlined in \refsec{sec:debugging-approach:implementation}.

\subsection{Idea}
\label{sec:debugging-approach:idea}
When developing a program, the user commonly uses a small instance to test it.
In order to verify the correctness of the results obtained from the ASP solver, the expected solution of the sample instance is determined by hand. That is, at least one answer set of an intended program for the given instance is known to the user.
A bug in the answer set program under test is then revealed when:
\begin{enumerate}
  \item[(a)] there are no answer sets (i.e. the program is incoherent), 
  or
  \item[(b)] the known answer set is not among the computed ones or there are answer sets corresponding to non-solutions of the sample instance.
\end{enumerate}

\begin{example}
[ASP conference system \cite{oetsch:2010:catching}]
\label{ex:bidding}
  The DLP $\pr'$ below models a conference system that assigns papers to program committee members.
  The assignment is done according to bids (ranging from $0$ to $3$) expressing a degree of preference on the papers.
  If no explicit bid is placed, a default value of 1 is assumed.
  \begin{align*}
    \pr' = \{ & pc(m_1), pc(m_2), paper(p_1), bid(m_1, p_1, 2), \\
             & some_-bid(M,P) \imp bid(M,P,X), \\
             & bid(M,P,1) \imp \dnot some_-bid(M, P), pc(M), paper(P) \}
  \end{align*}
  We expect a solution containing $bid(m_1,p_1,2)$ and $bid(m_2,p_1,1)$, but $\pr'$ is incoherent.
  Therefore, a bug of type (a) is revealed.
\end{example}

As illustrated by the example above, bugs of type (a) are revealed when the ASP solver finds that the program is incoherent.
In order to reveal a bug of type (b), additional information about the expected answer set is required.
This information is given in the form of a \emph{test case}, which intuitively asserts a set of literals to be true in some answer set of the faulty program $\Pi$.
Whenever there exists an answer set of $\Pi$ such that all asserted literals are true, the test case \emph{passes}.
If no such answer set exists, the test case \emph{fails} since no answer set models all asserted literals and a bug is revealed.

\begin{example}
  \label{ex:test-case}
  Consider a program $\pr''$ consisting of the following rules:
  \begin{align*}
    wet \lor dry & \imp \\
    umbrella \lor no\_umbrella & \imp \\
     & \imp wet,\; umbrella \\
     & \imp rainy,\; dry \\
     & \imp wet,\; \dnot rainy\\
    rainy &
  \end{align*}
  The user expects an answer set of $\pr''$ where $dry$ and $umbrella$ are true.
  However, $\pr''$ has only one answer set $\{ rainy, wet,  no\_umbrella\}$, thus, intuitively, we have a test case that fails.
\end{example}

\begin{definition}[Test Case]
  \label{def:test-case}
  A \emph{test case} for a program $\pr$ is a set of literals $T$ asserted to be true in some answer set.
  A test case \emph{fails}, if the program
  \begin{equation}
  \pr_T = \pr \cup \{ \imp \overline{l} \mid l \in T \}
  \end{equation}
  is incoherent.
\end{definition}

\begin{example}
  \label{ex:test-case-ctd}
  Consider the program $\pr''$ from Example~\ref{ex:test-case}.
  According to Definition~\ref{def:test-case}, the test case is represented by the set $T = \{ dry, umbrella \}$.
  The program $\pr''_{T}$ extends $\pr''$ by the constraints $\imp \dnot dry$, and $\imp \dnot umbrella$.
  Since $dry$ cannot be derived in $\pr''_{T}$, the program is incoherent.
\end{example}

We model assertions by constraints that forbid any answer set containing the complement of the asserted literals.
As a result, checking whether a test case $T$ of a program $\Pi$ passes or not is reduced to checking whether $\Pi_T$ is coherent, as illustrated in Example~\ref{ex:test-case-ctd}.
Hence the second case (b) of when a bug is revealed is reduced to the first case of incoherent programs.
Therefore, it is sufficient to focus on debugging of incoherent programs only.

Given an incoherent program $\pr$ and a test case $T$, the goal is to highlight a set of rules of $\pr_T$ that cause the incoherence.
Intuitively, not all rules of $\pr$ contribute to its incoherence.

\begin{example}
  Consider the program $\pr''$ and a test case $T = \{ dry, umbrella \}$ from Example~\ref{ex:test-case-ctd}.
  A debugger should identify the buggy rule:
  \begin{align*}
     & \imp rainy,\; dry \enspace.
  \end{align*}
  Indeed, given the rule $\imp \dnot dry$ and the fact $rainy$, the above constraint cannot be satisfied. In that case the buggy constraint should be replaced by:
  \begin{align*}
     & \imp rainy,\; dry, \; \dnot umbrella \enspace.
  \end{align*}
\end{example}

Unfortunately, the set of buggy rules might be large, thus not helping the user to localize the fault.
Therefore, in our approach the user is queried (in a smart and non-overwhelming way) to retrieve further information about the expected solution. Every query allows the debugger to exclude irrelevant rules and identify the buggy ones more precisely.
We implement this debugging strategy by using the concepts of solving under assumption and unsatisfiable cores \cite{zhang:2003:validating} as described in the next section. 

\subsection{The \dwasp Strategy}
\label{sec:debugging-approach:implementation}
In this section, we focus on how to integrate the debugging approach inside an ASP solver.
The task of debugging an incoherent answer set program is computationally hard.
Thus, integrating the approach inside an ASP solver aims to speed-up the debugging process.
For instance, the solving-under-assumptions feature \cite{een:2003:temporal} implemented in modern ASP solvers \cite{alviano:2015:advances,gebser:2012:answer} is used to compute an unsatisfiable core.
In a nutshell, assumptions correspond to a set of literals $A$ considered as true during the solving process.
Whenever some of the assumptions are violated during the solving process, the conflicting set of literals $C \subseteq A$, called \textit{unsatisfiable core}, is computed.

In order to utilize the solving-under-assumptions interface, a fresh \emph{debug atom} is introduced to the body of each rule of $\pr$, as defined in \refdef{def:debugging-program}.
Furthermore, users can specify a set of rules $\bg \subseteq \pr$ called \emph{background knowledge} that are considered to be correct. 
In the following, we assume $\bg$ to comprise all facts of a logic program $\pr$.

\begin{definition}[Debugging Program]
\label{def:debugging-program}
  Let $\pr$ be an incoherent DLP, $\bg$ be the background knowledge, and $id : (\pr \setminus \bg) \to \mathbb{N}$ be an assignment of unique identifiers to the non-background knowledge rules of $\pr$.
  Then, the \emph{debugging program} $\dpr{\pr}$ of $\pr$ with respect to the background knowledge $\bg$ is defined as
  \begin{equation}
  \begin{split}
  \dpr{\pr} = \{
  a_1 \lor \dots \lor a_m \imp  l_1, \dots,& l_n,  \debug{id(r)}{\vec{\mathit{vars}}} 
  \mid r \in (\pr \setminus \bg), \\
  & H(r) = \{a_1,\dots,a_m\}, B(r) = \{l_1,\dots,l_n\}\} \enspace,
  \end{split}
  \end{equation}
  where $\debug{id(r)}{\vec{\mathit{vars}}}$ is a fresh \emph{debug atom} and $\vec{\mathit{vars}}$ is a tuple comprising all variables of $B(r)$.
\end{definition}

\begin{example}
  \label{ex:bidding:ctd1}
  Consider the program $\pr'$ from Example~\ref{ex:bidding}.
  The corresponding debugging program is given as follows:
  \begin{align*}
    \dpr{\pr'} = \{ 
    & pc(m_1), pc(m_2), paper(p_1), bid(m_1, p_1, 2), \\
    & some_-bid(M,P) \imp bid(M,P,X), \debug{1}{M,P,X}, \\
    & bid(M,P,1) \imp \dnot some_-bid(M, P), pc(M), paper(P), \debug{2}{M,P} \}
  \end{align*}
\end{example}

\paragraph{Debugging incoherent programs.}
Algorithm~\ref{alg:debugging} depicts the implementation of the debugging strategy inside the solver.
We consider an incoherent program $\pr$ for debugging and input its ground debugging program ${\gdpr{\pr}}$ to the debugger.
First, we gather all debug atoms in the set $A$ (line \ref{alg:debugging:setA}).
Solving under the assumption that all debug atoms $A$ are true causes the solver to return a minimal unsatisfiable core $C$ containing debug atoms only (line \ref{alg:debugging:getC}).
Debug atoms with the same identifier $id_r$ correspond to the (non-ground) rule $r \in \pr$, while a ground debug atom corresponds to exactly one ground rule of $\pr$.
Thus, the atoms inside the minimized unsatisfiable core uniquely identify the set of ground and non-ground rules of $\pr$ that cause the incoherence.
We notify the user interface with these rules, which in turn highlights the rules to the user (line \ref{alg:debugging:notifyUI}).
Finally, we compute a query and issue it to the user, in order to add additional information to the set of assumptions $A$ (lines \ref{alg:debugging:queryStart}-\ref{alg:debugging:queryEnd}) and start a new debugging iteration.
\begin{algorithm}
	\caption{Debugging an incoherent logic program $\Pi$}
	\label{alg:debugging}
	\Input{A ground debugging program ${\gdpr{\Pi}}$}
	\Begin{
		$A := \{ d \mid \text{$d$ is a debug atom of ${\gdpr{\pr}}$} \}$ \; \label{alg:debugging:setA}
		\While{user continues debugging session}{
			$C := \text{compute minimal unsatisfiable core under assumptions $A$}$ \; \label{alg:debugging:getC}
			notify user interface with the rules corresponding to $C$ \; \label{alg:debugging:notifyUI}
			$q := \text{compute query atom using $C$ and $A$}$ \; \label{alg:debugging:queryStart}
			\eIf{user answers that $q$ is expected to be true}{
				$A := A \cup \{ q \}$ \;
			}{
			$A := A \cup \{ \dnot q \}$ \; \label{alg:debugging:queryEnd}
		}
	}
}
\end{algorithm}

\begin{example}
  \label{ex:bidding:ctd2}
  Consider the debugging program $\dpr{\pr'}$ from Example~\ref{ex:bidding:ctd1}.
  The ground instantiation of the debugging program is:
  \begin{align*}
    {\gdpr{\pr'}} = \{ 
      & pc(m_1), pc(m_2), paper(p_1), bid(m_1, p_1, 2), \\
      & some_-bid(m_1,p_1) \imp bid(m_1,p_1,1), \debug{1}{m_1,p_1,1}, \\
      & some_-bid(m_1,p_1) \imp bid(m_1,p_1,2), \debug{1}{m_1,p_1,2}, \\
      & some_-bid(m_2,p_1) \imp bid(m_2,p_1,1), \debug{1}{m_2,p_1,1}, \\
      & bid(m_1,p_1,1) \imp \dnot some_-bid(m_1, p_1), pc(m_1), paper(p_1),  \debug{2}{m_1,p_1}, \\
      & bid(m_2,p_1,1) \imp \dnot some_-bid(m_2, p_1), pc(m_2), paper(p_1), \debug{2}{m_2,p_1} \}
  \end{align*}
  In line \ref{alg:debugging:setA}, we add to the set of assumptions $A$ all debugging atoms:
  \begin{align*}
    A = \{ & \debug{1}{m_1,p_1,1}, \debug{1}{m_1,p_1,2}, \debug{1}{m_1,p_1,1}, \\
           & \debug{2}{m_1,p_1}, \debug{2}{m_2,p_1} \}
  \end{align*}
  The solver computes the minimal unsatisfiable core using $A$ in line \ref{alg:debugging:getC}:
  \begin{align*}
    C = \{ \debug{1}{m_2, p_1, 1}, \debug{2}{m_2, p1} \}
  \end{align*}
  The debugging atoms in $C$ correspond to the ground rules
  \begin{align*}
    & some_-bid(m_2,p_1) \imp bid(m_2,p_1,1), \debug{1}{m_2,p_1,1} \\
    & bid(m_2,p_1,1) \imp \dnot some_-bid(m_2, p_1), pc(m_2), paper(p_1), \debug{2}{m_2,p_1}
  \end{align*}
  which in turn correspond to the following non-ground rules of $\pr$:
  \begin{align*}
    & some_-bid(M,P) \imp bid(M,P,X), \\
    & bid(M,P,1) \imp \dnot some_-bid(M, P), pc(M), paper(P)
  \end{align*}
  In line \ref{alg:debugging:queryStart}, the debugger determines $q = bid(m_2,p_1,1)$ as query atom.
  As we expect a solution containing $bid(m_2, p_1, 1)$, we answer that $q$ is expected to be true, which causes the solver to extend the set of assumptions $A$ by $bid(m_2, p_1, 1)$.
  In the next iteration, a new unsatisfiable core $C$ is returned (line \ref{alg:debugging:getC}):
  \begin{align*}
    C = \{ \debug{1}{m_2,p_1,1} \}
  \end{align*}
  The core $C$ corresponds to following ground and non-ground rules
  \begin{align*}
    & some_-bid(m_2,p_1) \imp bid(m_2,p_1,1) \\
    & some_-bid(M,P) \imp bid(M,P,X)
  \end{align*}
  of the faulty program $\pr'$.
  We now see that the bug is caused by the outlined rule, as it derives $some_-bid(m_2,p_1)$ given $bid(m_2, p_1, 1)$, which in turn is derived as default by the last rules of $\pr'$.
\end{example}

\paragraph{Query computation.}
In order to narrow the source of the incoherence, queries are used, as pointed out in the previous section.
The computation of the queries is done by \emph{relaxing} the unsatisfiable core, that is by removing some debugging atom from the set of assumptions until an answer set is found.
A \emph{diagnosis} is a set of debug atoms such that when they are removed from the set of assumptions, the relaxed program is coherent.
The goal is to present the correct diagnosis to the user, however many diagnoses might exist.
Ideally, a query is asked in a way that, regardless the answer to the query, the number of diagnosis is cut in half.
Therefore, we choose the query atom as the atom $q$ occurring in a half of the answer sets of the relaxed programs.
If the user considers $q$ to be true in the expected answer set, $q$ is added to the assumptions and $\dnot q$ otherwise.

\paragraph{Missing support.}
Recall that an atom $u$ is unsupported w.r.t. an interpretation $I$, if no rule derives the atom.
Thus, if a supported atom $u$ is true in an interpretation $I$, then $I$ cannot be an answer set.
Consider the case when the user asserts an atom $u$ to be true in a test case $T$ of $\pr$, i.e. $u \in T$, and $u$ is unsupported in any answer set $\pr$.
The debugger will compute an unsatisfiable core of $\pr_T$ consisting of the rule $\imp \dnot u$ only.
However, when there is no assertion $\imp \dnot u$ available, the computed core will be empty.
Therefore, an additional way of detecting unsupported atoms is required.
We extend the debugging program $\dpr{\pr}$ by the set of rules
\begin{align*}
  \{ a \imp \support{a} \mid a \text{ is an atom of } \gpr \} \enspace,
\end{align*}
where $\support{a}$ is a fresh atom called \emph{supporting atom} of $a$.
Assuming that the supporting atoms are false does not alter the semantics of the program.
However, the solver will now include the supporting atoms inside the unsatisfiable core, allowing the identification of unfounded atoms inside the core.
Therefore, during the debugging process depicted in Algorithm~\ref{alg:debugging}, we extend the set of assumptions $A$ by the set of literals
\begin{align*}
  \{ \dnot s \mid s \text{ is a supporting atom of } {\gdpr{\pr}} \} \enspace.
\end{align*}

We now identify an unsupported atom $u \in \pr$ by having the atom $\support{u}$ inside an unsatisfiable core $C$ during the debugging process.
Intuitively, the fault is rooted in some rule $r$ failing to derive $u$, because (i) the body of $r$ is not satisfied, or (ii) the body of $r$ is satisfied but another atom of the head of $r$ is chosen.
Therefore, we select a query atom out of the set of atoms
\begin{equation}
  Q = \bigcup_{r \in \left\{ r \mid u \in H(r) \right\}}
    (H(r) \setminus \{ u \}) \cup 
    B^+(r) \cup
    \left\{ a \mid \dnot a \in B^-(r) \right\} \enspace.
\end{equation}

\section{The \dwasp System}
\label{sec:system-description}
In this section, we first give an overview of the debugging system and how the components interact with each other.
Afterwards, we describe the graphical user interface \dwaspgui in detail.
Furthermore, we present the communication protocol between the GUI and the debugger.
Finally, the integration with an integrated development environment for ASP, called \aspide \cite{febbraro:2011:aspide}, is presented.

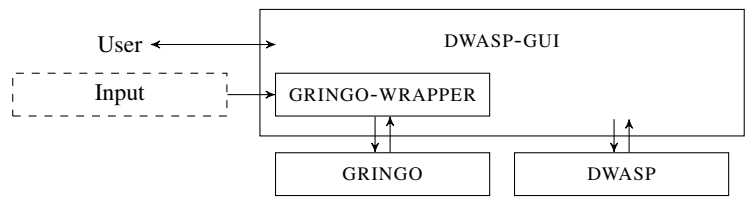
\begin{figure}
  \centering
\begin{tikzpicture}[->,>=stealth']
\node (user) {User};
\node[block, dashed, below=0.5em of user]   (input) {Input};
\node[block, right=2em of input]            (gringo-wrapper) {\gringowrapper};
\node[block, below=1.5em of gringo-wrapper] (gringo) {\gringo};
\node[block, right=1em of gringo]           (dwasp) {\dwasp};

\path (gringo) -- node[above=5em] (dwaspgui) {\dwaspgui} (dwasp);
\draw ($(gringo.north west)+(-0.2,1.9)$) rectangle ($(dwasp.south east)+(0.2,0.8)$);

\draw ($(gringo-wrapper.south)+(-.1,0)$) -> ($(gringo.north)+(-.1,0)$);
\draw ($(gringo.north)+(.1,0)$)          -> ($(gringo-wrapper.south)+(.1,0)$);
\draw (input)                            -> (gringo-wrapper);
\draw[<-] ($(dwasp.north)+(-.1,0)$)      -- +(0,0.44);
\draw[->] ($(dwasp.north)+(.1,0)$)       -- +(0,0.44);
\draw[<->] (user)                        -- +(2.07,0);
\end{tikzpicture}
  \caption{Interaction of the user with the debugging system: The front-end \dwaspgui uses \gringowrapper and \dwasp to debug the program.}
  \label{fig:architecture}
\end{figure}

Our system  consists of three components: \gringowrapper\ -- the debugging grounder, the solver \dwasp, and the graphical user interface \dwaspgui, as depicted in \reffig{fig:architecture}.
The user interacts with \dwaspgui and provides a program $\pr$ and some test case $T$.
If the test case fails, i.e. $\pr_T$ is incoherent, a new debugging session is started.
First, \gringowrapper transforms $\pr_T$ to the debugging program $\dpr{\pr_T}$.
Then, the debugging program is passed to \gringo%
\footnote{Note that simplifications of \gringo are disabled \cite{dodaro:2015:interactive}}
in order to obtained the ground debugging program $\gdpr{\pr_T}$.
Afterwards, the debugger \dwasp is started with $\gdpr{\pr_T}$ as input.
Unsatisfiable cores and queries are computed and displayed to the user, until the fault is localized, as described in \refsec{sec:debugging-approach}.

\subsection{User Interface \dwaspgui}
\label{sec:system-description:dwaspgui}
\begin{figure}
  \centering
  \includegraphics[width=.9\linewidth]{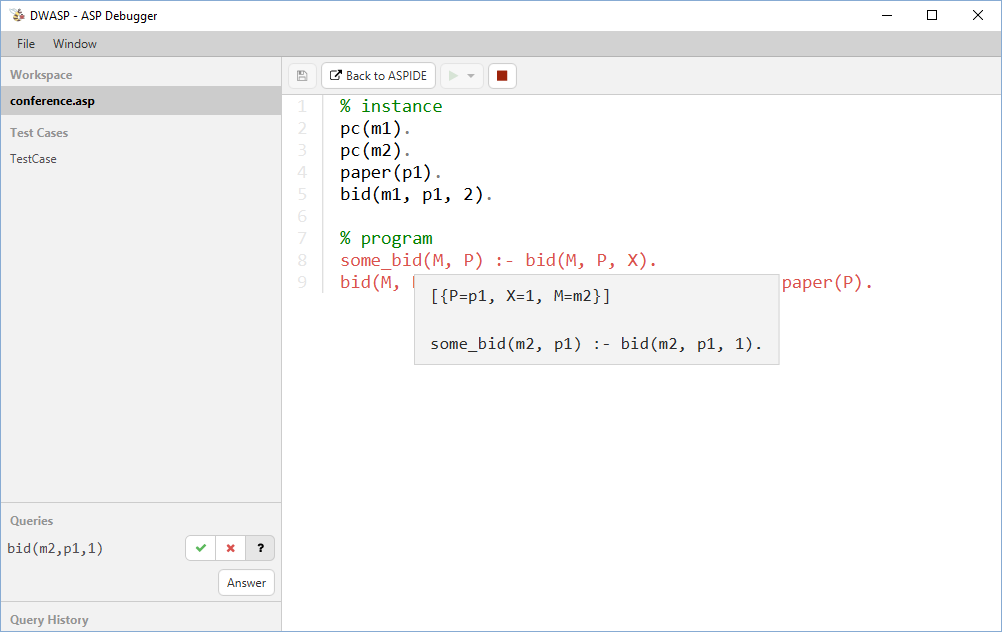}
  \caption{A screen-shot of \dwaspgui used to debug the program presented in Example~\ref{ex:bidding}.}
  \label{fig:dwasp-gui}
\end{figure}

A screenshot of \dwaspgui is depicted in \reffig{fig:dwasp-gui}.
The \emph{workspace}-view and \emph{test cases}-view list all files that contain the program encodings and test cases, respectively.
Furthermore, the \emph{queries}-view contains at most nine atoms (due to space restrictions of the GUI), whose truth-values are requested to be asserted by the user.
The user answers a query by selecting either the button with the check-mark or with the cross and clicking on \emph{send}.
Afterwards, \dwasp re-computes the unsatisfiable core and presents the results to the user.
While debugging a program, all rules that are contained in the current unsatisfiable core are highlighted in red.
When hovering over such a rule with the cursor, all substitutions as well as ground versions of the rules are displayed in a pop-up.

\subsection{Integration with \aspide}
\label{sec:system-description:aspide}
We integrated the graphical user interface \dwaspgui inside the integrated development environment \aspide \cite{febbraro:2011:aspide}.
The work-flow for testing and debugging is illustrated using the program presented in Example~\ref{ex:bidding}.

\begin{figure}
  \centering
  \includegraphics[width=.9\linewidth]{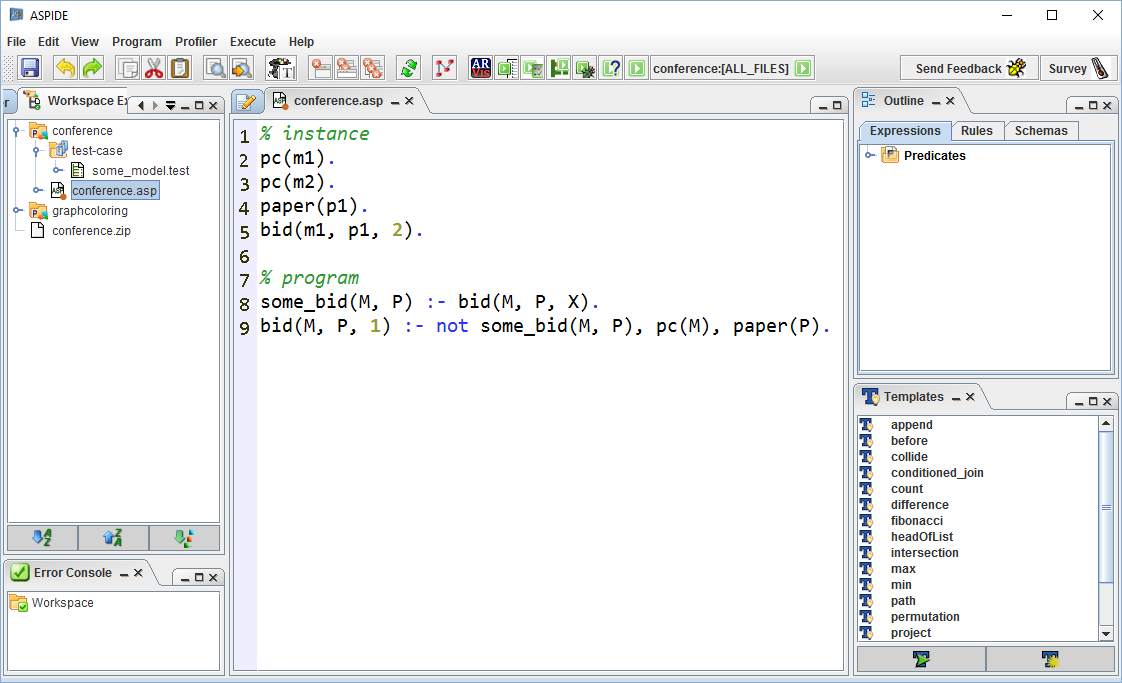}
  \caption{A screen-shot of \aspide displaying the program presented in Example~\ref{ex:bidding}.}
  \label{fig:aspide-i}
\end{figure}

\begin{figure}
  \centering
  \includegraphics[width=.9\linewidth]{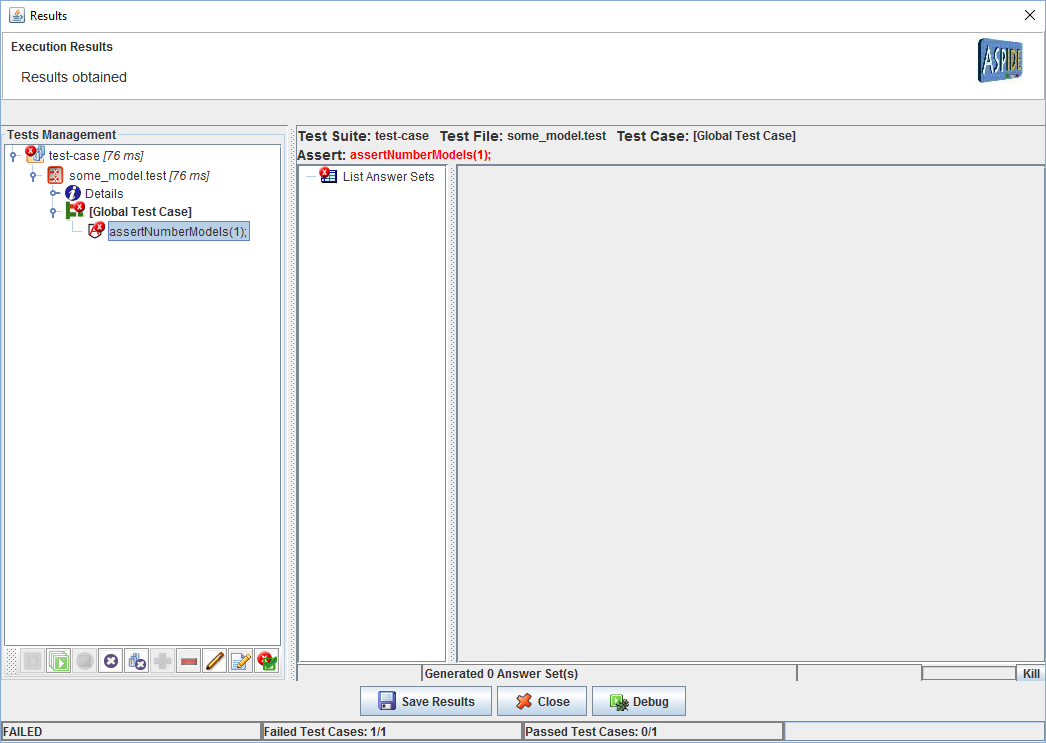}
  \caption{A screen-shot of \aspide displaying the failed test case.}
  \label{fig:aspide-ii}
\end{figure}

\begin{figure}
	\centering
	\includegraphics[width=.9\linewidth]{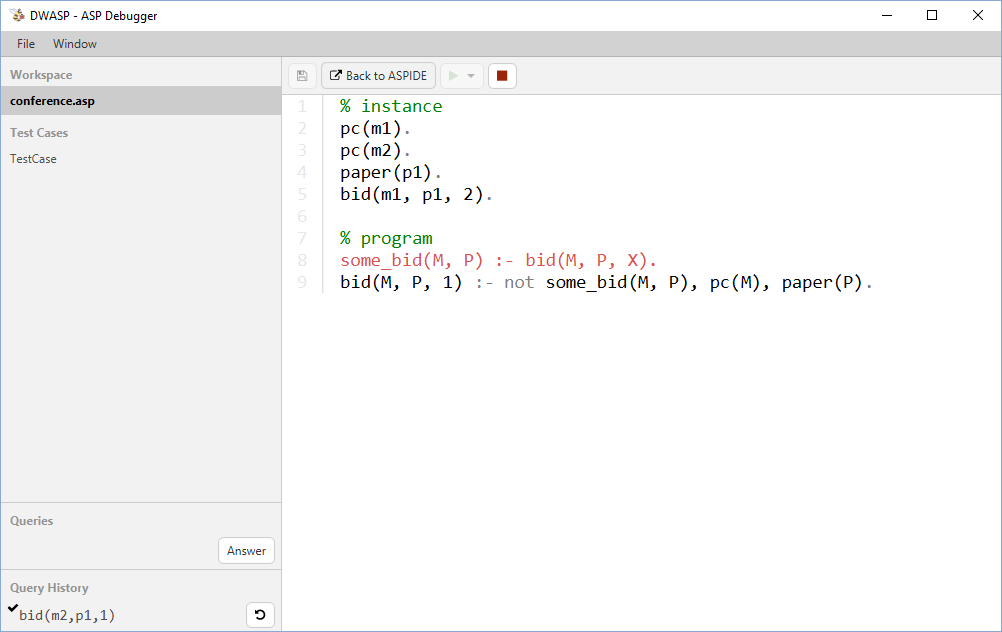}
	\caption{A screen-shot of \dwaspgui, were the faulty program is being debugged.}
	\label{fig:aspide-iii}
\end{figure}

In Fig. \ref{fig:aspide-i}, we present a screen-shot of \aspide with a workspace that has the program $\pr'$ loaded.
The test case \verb|some_model.test| uses an assertion of \aspide \cite{febbraro:2011:unit} that checks whether some answer set exists.
On executing the test case \verb|some_model.test|, the IDE tells us that the test case failed, as depicted in figure~\ref{fig:aspide-ii}.
In order to start the \dwasp-system, we click on the \emph{Debug} button.
We are now presented with the interface \dwaspgui as shown in figure~\ref{fig:aspide-iii}, where we debug the faulty program as described in the previous section.
Finally, we click on the \emph{Back to ASPIDE} button, which returns us to \aspide, having the faulty rule highlighted as well.

\section{Related Work}
\label{sec:related}

Modern ASP debugging approaches can be mainly separated into integrated and declarative approaches. 
The first approaches are based on a tight integration with the solver, whereas the second ones are solver independent and are based on \emph{meta-programming}.

The \textsc{dlv} debugger developed in~\cite{Perri2007Sea} is an example of an integrated approach. It uses the reason calculus 
to detect and store the choices made by the solver during the backtracking phases in a \textit{reasons table}. 
The table can be queried to justify the presence/absence of a literal in an answer set or to explain the incoherency of the program. 
This debugging system is however very limited, since it uses specific features of the \textsc{dlv} system and can only provide a partial interpretation justifying the lack of a model. 
%
%
\textsc{ideas}~\cite{Brain2005} is another procedural approach aiming at two types of problems: (a) why a set of atoms $S$ is in an answer set $M$ and (b) why $S$ is not in any answer set. 
Both \textsc{ideas} algorithms are similar to the ones implemented in ASP solvers and try to decide which rules are responsible for derivation or non-derivation of atoms in $S$. 
The interactivity of \textsc{ideas}, as well as of all other modern debuggers allows a programmer: (1) to query a system for an explanation of an observed fault, (2) analyze the obtained results and (3) reformulate the query to make it more precise. In our approach we reuse the algorithms implemented in a solver and are able to find required refinements automatically, thus, making the steps (2) and (3) obsolete.

The declarative debuggers use a program over a meta language -- a kind of ASP solver simulation -- to manipulate a program over an object language -- the faulty program. 
Each answer set of a meta-program comprises a \emph{diagnosis}, which is a set of meta-atoms describing the cause why some interpretation of the faulty program is not its answer set. 
An approach used in \textsc{smdebug}~\cite{Syrjanen2006} addresses debugging of incoherent non-disjunctive ASP programs by adaption of model-based diagnosis~\cite{Reiter87}. 
Similarly to our approach the debugger focuses on detection of odd loops, but cannot detect problems arising due to unfounded sets. The \textsc{spock}~\cite{Gebser2008b} and \textsc{Ouroboros}~\cite{Oetsch2010a,PolleresFSF13} debuggers extend \textsc{smdebug} by enabling identification of problems connected with unfounded sets. Both approaches represent the input program in a reified form allowing application of a debugging meta-program. In case of \textsc{spock} the debugging can be applied only to grounded programs, whereas \textsc{Ouroboros} can tackle non-grounded programs as well. The main problem of meta-programming approaches is that often the grounding of the debugging meta-program explodes. This is due to the fact that the ground debugging program has to comprise all atoms explaining all possible faults in an input faulty program, which is not the case in our approach. Moreover, our approach generalizes the interactive query-based method built on top of \textsc{spock}~\cite{Shchekotykhin15} by enabling its application to non-ground programs.

There are other approaches enabling faults localization in ASP, but not directly comparable with \dwasp, include Consistency-Restoring Prolog~\cite{Balduccini2003}, translation of ASP programs to natural language~\cite{Mikitiuk2007}, visualization of justifications for an answer set~~\cite{Pontelli2009} as well as stepping thought an ASP program~\cite{Oetsch2011step}. Combining these approaches with ideas implemented in \dwasp is a part of our future work.

In \cite{IDPIDE} a web-based programming environment for the IDP system is presented. The IDE also provides a graphical interface for a debugging approach based on assumptions and core-detection. However, \cite{IDPIDE} applies it to a different language and it does feature a question-answering schema that is fundamental for reducing the set of buggy rules.

In \cite{tingting:2015:ilp}, the authors present a debugging technique for normal ASP programs that is based on inductive logic programming (ILP) and test cases. The idea is to allow the programmer to specify test cases modeling features that are expected to appear in some solution and those that should not. 
These are used to to revise the original program semi-automatically so that it satisfies the stated properties.
The implementation of the theory revision is done in ASP using an abductive logic programming technique.
This approach can complement our debugging approach since it has the possibility to learn rules (and modifications of rules), whereas we focus on finding errors assuming the program is a complete specification.

In \cite{schulz:2015:lpnmrdebug} a different approach to debugging ASP programs is presented, and the reason of an incoherence is studied in terms of a set of culprits (atoms) using semantics which are weaker than the answer set semantics. They also provide a technique for explaining the set of culprits in terms of derivations. This idea is further extended in \cite{schulz:2016:argumentation}, where argumentation theory is used to explain why a literal is or is not contained in a given answer set, and providing a means for studying relationships among literals. These approaches see the reason of a bug in the truth of a set of atoms, thus are, in a sense, complementary to our approach (we identify the rules involved in a conflict).

\section{Conclusion}
\label{sec:conclusion}
In this paper a new graphical interface for the \dwasp debugger has been presented.
The new interface improves the user-experience of debugging ASP programs with \dwasp, a process that was possible before only trough a command line interface. 
Indeed, besides the usual advantages provided by visual tools, the new interface simplifies two tasks that are not easy to carry out in the command line interface, namely: the definition of test cases and the interactive query answering. The query answering feature is much more user friendly, since the user can simply select answers by clicking on dedicated buttons, and several possible answers are presented to the user in a convenient list.
Several test cases can be easily loaded in the interface, and several debugging session can be seamlessly run on the same cases if needed. Also problematic rules are outlined immediately in the text editor so the user is pointed immediately from the interface to sources of bugs.
\dwaspgui has also been integrated in \aspide, which was missing a complete debugger interface supporting non ground ASP programs. The integration includes specific support for creating failing test cases to debug directly from the unit test framework provided by \aspide.
With our extension \aspide turns into a full-fledged IDE by offering complete debugging support.

\paragraph*{Acknowledgments.}
The authors are grateful to Marc Deneker and Ingmar Dasseville for the fruitful discussions about debugging ASP programs, and in particular for the useful comments regarding the case of missing support. 

This work was partially supported by the Austrian Science Fund (FWF) contract number I 2144 N-15, the Carinthian Science Fund (KWF) contract KWF-3520/26767/38701, the Italian Ministry of University, Research under PON project ``Ba2Know (Business Analytics to Know) Service Innovation -- LAB'', No. \ PON03PE\_00001\_1, and by the Italian Ministry of Economic Development under project ``PIUCultura (Paradigmi Innovativi per l'Utilizzo della Cultura)'' n.\ F/020016/01--02/X27.

\bibliographystyle{splncs03}
\bibliography{paper}
\end{document}